\documentclass[10pt,twocolumn,letterpaper]{article}

\usepackage[pagenumbers]{iccv} % To force page numbers, e.g. for an arXiv version
\usepackage{multicol}
\usepackage{multirow}
\usepackage[accsupp]{axessibility}  % Improves PDF readability for those with disabilities.
\newcommand{\ours}{BabyVLM\xspace}

\newcommand{\myparagraph}[1]{\smallskip\noindent\textbf{#1}}

\usepackage{makecell}
\usepackage{pifont}
\usepackage{afterpage}
\usepackage{tabularx}
\usepackage{float}

\definecolor{iccvblue}{rgb}{0.21,0.49,0.74}
\usepackage[pagebackref,breaklinks,colorlinks,allcolors=iccvblue]{hyperref}

\def\paperID{10281} % *** Enter the Paper ID here
\def\confName{ICCV}
\def\confYear{2025}

\title{\ours: Data-Efficient Pretraining of VLMs Inspired by Infant Learning\thanks{\href{https://shawnking98.github.io/BabyVLM/}{Project website: \texttt{shawnking98.github.io/BabyVLM}}}}

\author{
\begin{tabular}{ccc}
Shengao Wang & Arjun Chandra & Aoming Liu \\
Boston University & Boston University & Boston University \\
{\tt\small wsashawn@bu.edu} & {\tt\small ac25@bu.edu} & {\tt\small amliu@bu.edu} \\
\end{tabular}
\and
\begin{tabular}{cc}
Venkatesh Saligrama & Boqing Gong \\
Boston University & Boston University \\
{\tt\small srv@bu.edu} & {\tt\small bgong@bu.edu} \\
\end{tabular}
}

\begin{document}
\maketitle

%I think the images for the benchmarks should be bigger here

\begin{figure*}[!h]
    \centering
    \includegraphics[width=1.\linewidth]{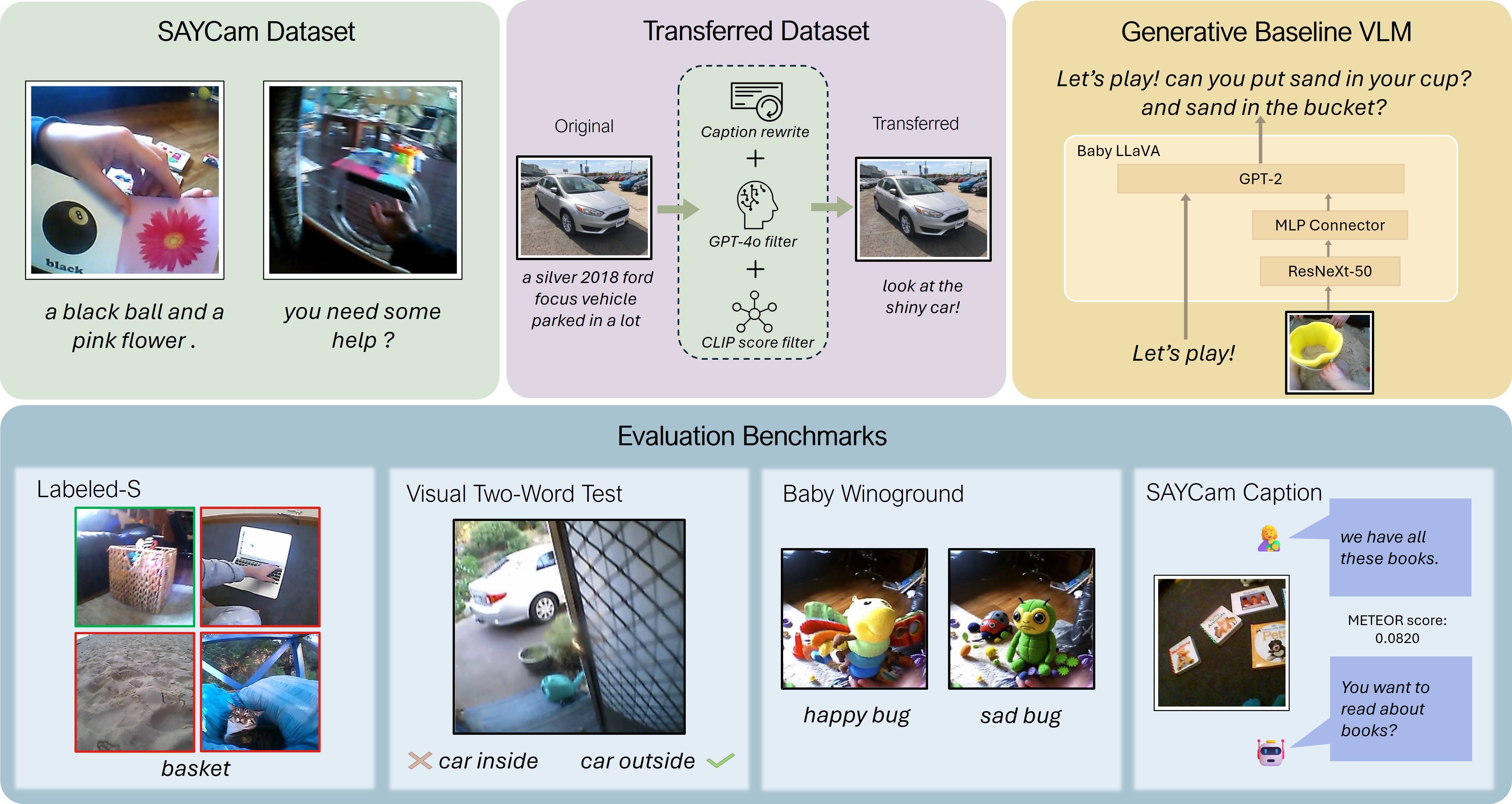}
    \caption{We introduce \ours, a developmentally inspired framework derived from SAYCam, consisting of the original SAYCam dataset \cite{sullivan2021saycam}, a transferred training dataset, a generative baseline VLM, and four evaluation benchmarks.}
    \label{fig:SAYCamBench overview}
\end{figure*}

\begin{abstract}
Human infants rapidly develop visual reasoning skills from minimal input, suggesting that developmentally inspired pretraining could significantly enhance the efficiency of vision-language models (VLMs). Although recent efforts have leveraged infant-inspired datasets like SAYCam, existing evaluation benchmarks remain misaligned—they are either too simplistic, narrowly scoped, or tailored for large-scale pretrained models. Additionally, training exclusively on SAYCam overlooks the broader, diverse input from which infants naturally learn. To address these limitations, we propose \ours, a novel framework comprising diverse in-domain evaluation benchmarks and a synthetic training dataset created via child-directed transformations of existing datasets. We demonstrate that VLMs trained with our synthetic dataset achieve superior performance on \ours{} tasks compared to models trained solely on SAYCam or general-purpose data of the \mbox{SAYCam} size. \ours thus provides a robust, developmentally aligned evaluation tool and illustrates how compact models trained on carefully curated data can generalize effectively, opening pathways toward data-efficient vision-language learning paradigms.
\end{abstract}

\section{Introduction}
\begin{table*}[h]
    \centering
    \begin{tabular}{lccc} % No vertical lines
    \toprule
    \textbf{Benchmark} & \textbf{Task Diversity} & \textbf{Baby-like} & \textbf{In-domain} \\
    \midrule
    \makecell{General purpose (VQA \cite{VQA}, Winoground \cite{thrush2022winoground}, etc.)} & \ding{51} & \ding{55} & \ding{55} \\
    DevBench \cite{tan2024devbench} & \ding{51} & \ding{51} & \ding{55} \\
    Labeled-S \cite{orhan2020self} & \ding{55} & \ding{51} & \ding{51} \\
    ModelVsBaby \cite{sheybani2024modelvbaby} & \ding{55} & \ding{51} & \ding{55} \\
    MEWL \cite{jiang2023mewlfewshotmultimodalword} & \ding{51} & \ding{51} & \ding{55} \\
    \midrule
    \ours & \ding{51} & \ding{51} & \ding{51} \\
    \bottomrule
    \end{tabular}

    \caption{Representative features of existing multimodal evaluation benchmarks. \textbf{Task Diversity:} The benchmark should include diverse tasks that assess different aspects of a vision-language model’s capability rather than focusing solely on simple tasks (e.g., object classification). \textbf{Baby-like:} The benchmark should align with cognitive and linguistic developmental stages observed in human babies. \textbf{In-domain:} The testing samples should come from the same data domain as the training dataset, ensuring that evaluation results reflect the model’s ability to generalize within a realistic learning environment.}
    \label{tab:related benchmark}
\end{table*}

We propose a novel framework, \ours, for data-efficient pretraining of vision-language models (VLMs). To this end we introduce methods for creating minimal yet naturalistic data—akin to the input human infants receive—as well as diverse in-domain evaluation benchmarks. By carefully curating the training data, we show that our method yields more robust, baby-like representations compared to training on general-purpose corpora, and can further serve as a template for resource-efficient model training in other specialized domains.

\myparagraph{Challenges in Current VLM Training.} Vision-Language Models have advanced rapidly in recent years~\cite{CLIP,gemini,gpt4o,qwen2.5,llava}, but these advancements often rely on massive datasets and prohibitively expensive computational resources. For instance, training large-scale models such as LLaMA~\cite{touvron2023llama} or LLaVA~\cite{llava} can require thousands of GPU hours \cite{hoffmann2022training,strubell2019energy}. Such demands pose fundamental barriers for independent researchers with limited resources, highlighting the need for more accessible pretraining methods.

\myparagraph{Lessons from Infant Learning.} Human infants, by contrast, rapidly acquire complex cognitive and perceptual skills from minimal data and limited environmental exposure \cite{smith2017developmental,latourrette2023principled}. This exceptional efficiency implies that robust representations can be learned from small, carefully curated datasets when these datasets closely mimic natural developmental conditions. Recognizing this, researchers have begun curating datasets such as SAYCam~\cite{sullivan2021saycam}, which provides egocentric audiovisual recordings of infants aged 6–32 months. Although our work primarily utilizes SAYCam, other developmentally inspired datasets such as BabyView \cite{long2024babyview} also support this approach. Our framework capitalizes on these insights, suggesting that intentionally constrained, naturalistic training scenarios can yield efficient, highly generalizable models.

\myparagraph{The Evaluation Gap.} Despite the promise of data-efficient VLM training inspired by infant learning, evaluating such compact models remains a critical challenge. Current benchmarks—such as VQA \cite{VQA}, Winoground \cite{thrush2022winoground}, and COCO \cite{COCO}—were designed for large-scale models trained on massive datasets, assessing capabilities that exceed those reasonably achievable by developmentally plausible, compact models. For instance, the Labeled-S benchmark \cite{orhan2020self}, which specifically targets SAYCam data, evaluates only a single classification task and thus cannot comprehensively measure broader vision-language capabilities. Conversely, developmental psychology benchmarks \cite{tan2024devbench,orhan2020self,konkle2010conceptual} tend to be overly simplistic or not directly relevant to the infant-inspired training data. As summarized in Table~\ref{tab:related benchmark}, this evaluation gap underscores the need for comprehensive, developmentally aligned benchmarks—precisely the gap addressed by our proposed framework, \ours.

To bridge this evaluation gap and realize our goal of data-efficient, developmentally aligned VLM pretraining, we offer three main contributions:

\begin{itemize}
    \item \textbf{In-Domain Evaluation Tasks.} We design three novel evaluation tasks derived from the SAYCam dataset. These tasks are tailored to reflect the cognitive and perceptual abilities typical of early human development, enabling diverse and meaningful evaluation of compact models trained on developmentally plausible data.
    
    \item \textbf{Synthetic Data Augmentation.} We introduce a data distillation approach to address the inherent limitations of existing small-scale datasets. By synthesizing simplified, child-directed versions of existing datasets like CC3M~\cite{CC3M} using GPT-4o \cite{gpt4o}, we create training data that more closely mirrors the linguistic and visual complexity encountered by infants.
    
    \item \textbf{BabyLLaVA: Generative Model Trained from Scratch.} Inspired by recent methods \cite{vong2024cvcl,llava}, we present BabyLLaVA, the first generative VLM trained entirely on developmentally plausible data. BabyLLaVA demonstrates that compact generative models, when trained on intentionally constrained and naturalistic data, can produce robust, baby-oriented responses from the input of baby viewpoints.
\end{itemize}
% We demonstrate that the models pretrained on our synthetic dataset yields superior performance on \ours benchmarks compared to training exclusively on the original SAYCam data or on existing general-purpose datasets.

Collectively, these contributions not only demonstrate effective, resource-efficient pretraining within our specific domain but also offer insights that can inform efficient paradigms across diverse applications, thereby lowering barriers to foundational model research.

 %We believe that \ours offers the research community a robust tool for the nuanced evaluation of early-stage vision-language models, fostering advancements that closely align with human developmental processes.

\begin{figure*}[htbp]
    \centering
    \includegraphics[width=0.85\linewidth]{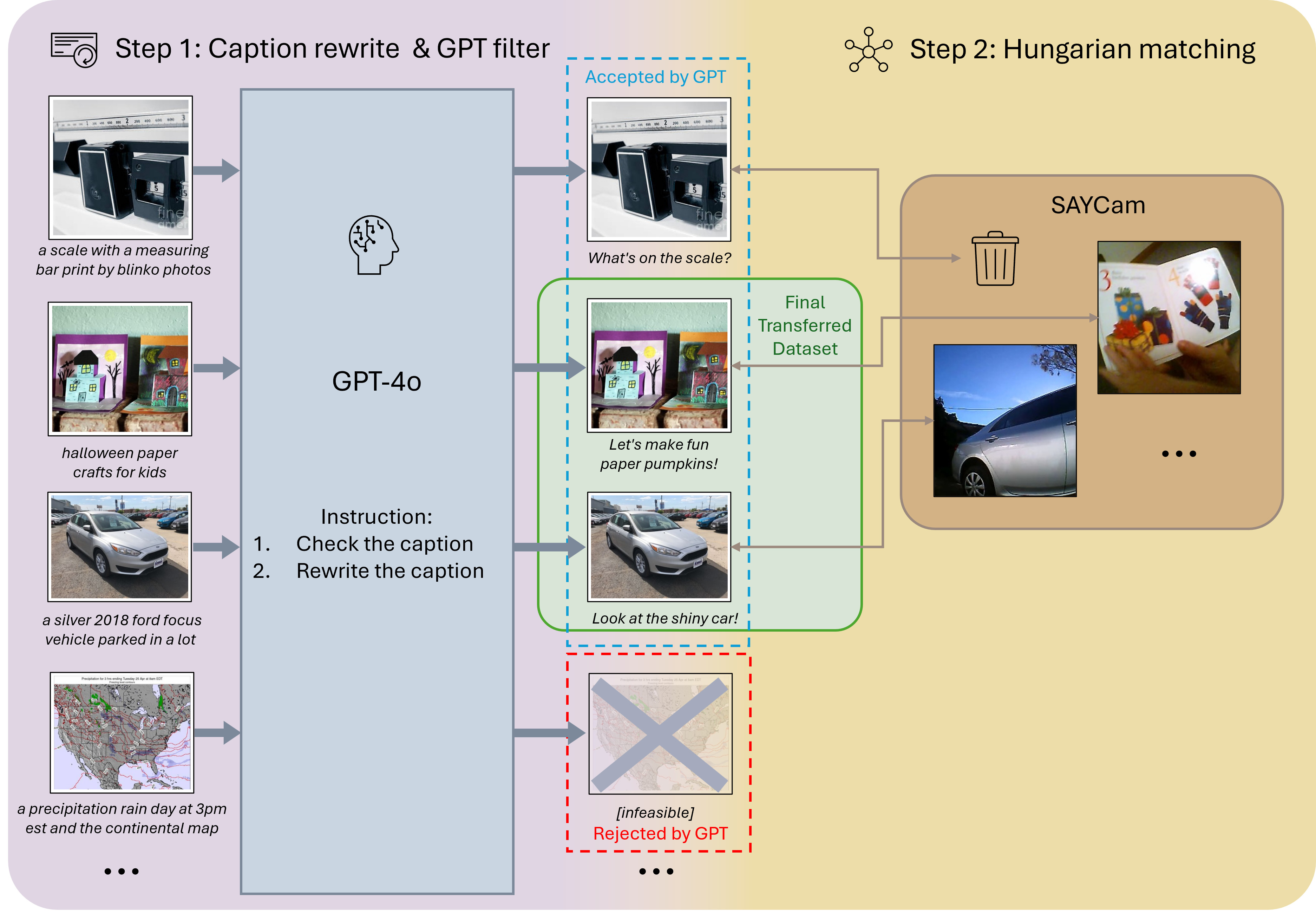}
    \caption{Pipeline for generating the transferred dataset. \textbf{Step 1:} We prompt GPT-4o to check whether an input caption is describing something a child would see in daily life and transfer the original image captions into simpler, child-directed utterances. \textbf{Step 2:} We use the CLIP similarity score as a metric to represent the distance between two images, and then conduct Hungarian matching to select a small subset of the transferred dataset that is visually aligned with SAYCam images.}
    \label{fig:transferred data pipeline}
\end{figure*}

\section{Related Work}

\textbf{Vision-Language Models.} Large vision-language models (VLMs)~\cite{CLIP,llava,gpt4o,gemini, bai2025qwen2, lu2024deepseek} have significantly advanced multimodal understanding by integrating visual and linguistic data for various tasks, including image captioning, visual question answering, and conversational interaction. Early influential models such as CLIP~\cite{CLIP} leveraged contrastive learning paradigms, effectively aligning visual and textual representations within a unified embedding space. More recent generative frameworks, such as LLaVA series~\cite{llava, llava1.5, liu2024llavanext, li2024llavanext-strong, zhang2024llavanext-video, li2024llavanext-ablations, li2024llavanext-interleave}, have combined pretrained visual encoders~\cite{CLIP, zhai2023siglip} with large language models~\cite{vicuna, bai2023qwen, team2024gemma, jiang2023mistral7b}, enabling more advanced conversational interactions and multimodal generative capabilities. However, these models typically require extensive computational resources and large-scale datasets. In contrast, our approach specifically targets compact generative VLMs trained exclusively on developmentally plausible datasets, providing a framework to improve data efficiency and better align model training with cognitive development processes observed in human infants.

\myparagraph{Developmentally Inspired Learning.} Human infants exhibit remarkable efficiency in acquiring language and visual concepts from limited and naturalistic input, inspiring substantial research into developmentally plausible training paradigms. Early influential datasets like CHILDES~\cite{macwhinney1990childes} facilitated initial explorations into language acquisition through linguistic recordings across diverse languages~\cite{Brent1996, alhama-etal-2020-evaluating}. Recent initiatives, including the BabyLM Challenge~\cite{choshen2024babylmchallenge, warstadt2023babylmchallenge}, further encouraged the development of models trained on language data scales comparable to those encountered by infants. Extending these ideas into multimodal contexts, datasets such as SAYCam~\cite{sullivan2021saycam} and BabyView~\cite{long2024babyview} have provided egocentric audiovisual data, enabling research that progresses from single-modality learning~\cite{orhan2024learning, sheybani2023curriculum, orhan2020self, wang2023finding, qin2024systematic} to visually grounded language acquisition~\cite{Zhuang2021UnsupervisedNN, vong2024cvcl}. Our work distinctly builds upon these foundations by explicitly creating synthetic, child-directed multimodal data from general-domain sources, addressing the limitations inherent in existing infant-inspired datasets and exploring the potential of compact generative VLMs trained in developmentally realistic conditions.

\myparagraph{Multimodal Benchmarks.} Existing multimodal evaluation benchmarks can be broadly classified as general-purpose or developmentally inspired. General-purpose benchmarks, such as Visual Question Answering (VQA) \cite{VQA, zhang2016yin} and Winoground \cite{thrush2022winoground}, evaluate advanced visio-linguistic integration but typically rely on large-scale, non-developmental datasets, rendering them unsuitable for compact models trained on limited developmental data. Conversely, developmentally inspired benchmarks—such as Labeled-S \cite{orhan2020self}, ModelVsBaby \cite{sheybani2024modelvbaby}, DevBench \cite{tan2024devbench}, and MEWL \cite{jiang2023mewlfewshotmultimodalword}—are more aligned with early cognitive processes but often limited to simplistic classification tasks or utilize data not fully reflective of the training domain. Our work explicitly addresses these gaps by proposing cognitively nuanced benchmarks directly aligned with the developmental data domain, thereby enabling accurate and relevant assessments of compact vision-language models trained from minimal, developmentally appropriate multimodal inputs.

\section{Framework}

Our proposed framework, \ours, aims to facilitate resource-efficient pretraining and developmentally aligned evaluation of compact VLMs inspired by the minimal yet highly informative learning environments of human infants. To achieve this, \ours{} comprises: (1) a filtered subset of baby-egocentric audio-visual recording from the SAYCam dataset, (2) a novel synthetic training dataset specifically crafted to reflect infant-directed linguistic and visual experiences, (3) a generative baseline model, BabyLLaVA, trained entirely on this developmentally plausible data, and (4) three novel evaluation benchmarks explicitly tailored to assess multimodal reasoning aligned with early cognitive stages, plus Labeled-S~\cite{orhan2020self}, an existing classification benchmark. Please refer to Figure~\ref{fig:SAYCamBench overview} for an overview of our framework.

\myparagraph{Design Principles for the \ours Framework.}
A central goal of \ours{} is to ensure realistic alignment with developmental constraints characteristic of early visual-language learning. To this end, we adopt the following concise guiding principles:
\begin{itemize}
    \item \textbf{Developmentally Appropriate Complexity:} 
    Tasks reflect cognitive capabilities typical of early developmental stages (e.g., basic object and action recognition, simple compositional reasoning), explicitly avoiding tasks requiring more complex reasoning.

    \item \textbf{Limited Generalization Beyond Early Development:} 
    Models should demonstrate intentionally constrained generalization, ignoring performance beyond realistic developmental boundaries.

    \item \textbf{Linguistic and Visual Simplicity:} 
    Dataset construction explicitly emphasizes simple vocabulary, concrete visual scenes, and straightforward grammatical structures consistent with child-directed interactions.
\end{itemize}

These principles collectively ensure that the resulting BabyVLMs remain authentic representations of early-stage developmental models, with their effectiveness empirically confirmed in our out-of-domain task evaluations, which are presented in the supplementary material.

\subsection{Datasets}
\noindent \textbf{Filtered SAYCam Dataset.} \label{sec:filtered_saycam_dataset}
% The original SAYCam dataset \cite{sullivan2021saycam} consists of egocentric audiovisual recordings. Following prior preprocessing by Vong et al.~\cite{vong2024cvcl}, we extract only child-directed speech and sample video into image-utterance pairs. To further align this dataset with developmental appropriateness, we refined the corpus by calculating CLIP similarity scores \cite{CLIP} between each image and its associated caption, retaining only pairs exceeding an experimentally determined similarity threshold of 0.2. This filtering ensures both linguistic and visual simplicity, consistent with the minimal complexity of early developmental stages, and results in approximately 67K image-utterance pairs. Examples of the filtered SAYCam dataset are provided in the supplementary material.
The original SAYCam dataset~\cite{sullivan2021saycam} comprises egocentric audiovisual recordings. Following Vong et al.~\cite{vong2024cvcl}, we extract child-directed speech and sample video into image-utterance pairs. To improve data quality, we filter pairs using CLIP similarities~\cite{CLIP}, retaining only those above a 0.2 threshold to ensure high image-text relevance. This results in ~67K pairs. Examples are provided in the supplementary material.

% \subsection{Filtered SAYCam Dataset} \label{sec:filtered_saycam_dataset}
% The original SAYCam dataset \cite{sullivan2021saycam} consists of egocentric audiovisual recordings. Following prior preprocessing by Vong et al.~\cite{vong2024cvcl}, we extract only child-directed speech and sample video into image-utterance pairs. To improve the quality of the dataset, we further refined this corpus by calculating CLIP similarity scores \cite{CLIP} between each image and its associated caption, retaining only pairs exceeding an experimentally determined similarity threshold of 0.2. Examples of the filtered SAYCam dataset are provided in the supplementary material.

% I think we can frame this section differently. We introduce augmented data from LLaVA pretraining not because of dataset distillation but rather because baby's learning signal is much richer than image caption pairs that we currently use. We can potentially combine this section with the Developmentally-inspired learning section and cite preivous literature (e.g. https://arxiv.org/pdf/2009.08497, https://www.nature.com/articles/s42256-022-00488-2) to make two claims (1) Babies have rich and diverse learning signals and (2) Other works in DIL also leverage unsupervised learning, data augmentation, or other techniques to mimic this. That will position our synthetic dataset better wtihin the literature than comparing it to data distillation.  This probably doesn't need to be mentioned in related works though, we can motivate it later when we introduce ?

\begin{figure*}[htbp]
    \centering
    \includegraphics[width=1\linewidth]{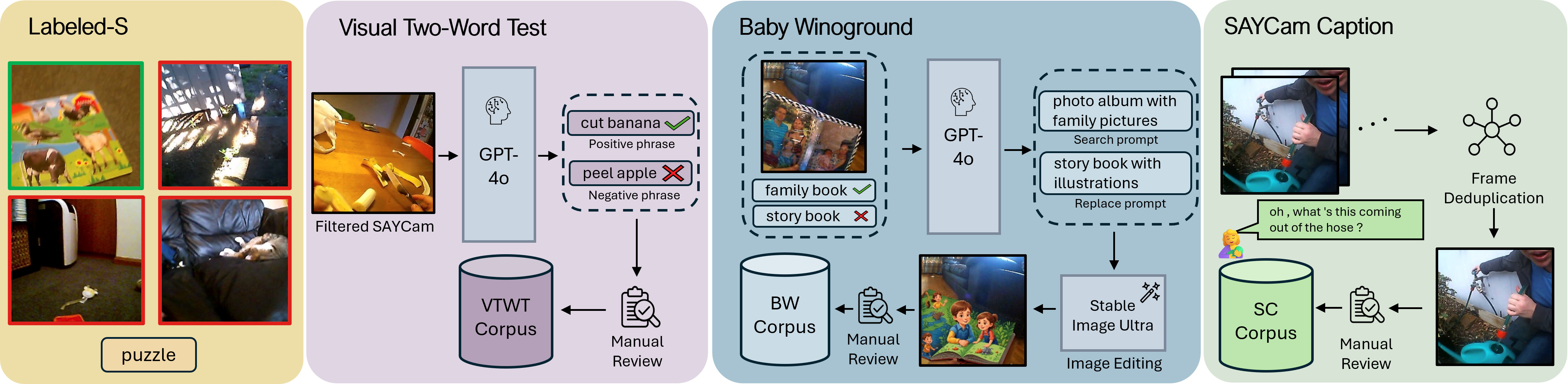}
    \caption{Illustrations of in-domain evaluation benchmarks in the \ours framework. \textbf{Labeled-S:} The category label must be matched to the target referent among 4 candidates. \textbf{Visual Two-Word Test:} The positive phrase must be matched to the image. Positive and negative phrases are generated by GPT-4o. \textbf{Baby Winoground:} The positive and negative phrases must be matched with their corresponding images. Negative images are generated by Stable Diffusion~\cite{stablediffusion3}, with prompts enhanced by GPT-4o. \textbf{SAYCam Caption:} The generated image caption must match the ground truth image caption. All image-caption pairs come from a de-duplicated subset of the SAYCam test split.}
    \label{fig: benchmark pipeline}
\end{figure*}

\myparagraph{Transferred Synthetic Training Dataset.}
\label{sec:transferred dataset}
While the SAYCam dataset provides naturally curated developmental data, there are inherent limitations of relying on this dataset exclusively. First, videos in SAYCam were recorded in 60 to 80 minute sessions twice a week, documenting only a small subset of each child's developmental experience. Moreover, due to practical constraints, SAYCam videos were often recorded at fixed times each week, reducing variation in the infant's recorded environment and further limiting our ability to capture the diverse multimodal input streams from which babies learn \cite{zaadnoordijk2020bigthingsunsupervisedmachine}.
%Actually we don't address temporal info loss so maybe not worth adding this 
%Moreover, by preprocessing videos to generate image-utterance pairs, we discard temporal information, which is a crucial learning signal during the early stages of human infant development \cite{marcovitch2009sequence}.
To address these limitations, we created a synthetic auxiliary training corpus by adapting general-purpose multimodal datasets—CC3M~\cite{CC3M}, LAION~\cite{schuhmann2022laion}, and SBU~\cite{ordonez2011im2text}—to match infant learning conditions.

Our approach comprises two steps, as illustrated in \mbox{Figure}~\ref{fig:transferred data pipeline}. In the first step, we prompt GPT-4o to rewrite captions into concise, child-directed utterances, emulating speech typically used with two-year-olds. GPT-4o also flags which image-caption pairs are misaligned with an infant's daily experience for exclusion. By emphasizing everyday vocabulary, simple grammar, and concrete objects and actions, we ensured linguistic alignment with early-stage learners. 

In the second step, to further maintain visual consistency, we apply the Hungarian algorithm \cite{jonker1988shortest} and utilize the CLIP similarity as a distance metric to select a subset of images resembing SAYCam. The number of selected samples matches that of the filtered SAYCam training set, resulting in a dataset that maintains visual alignment while balancing diversity and domain relevance. More details of the transformation guidelines and examples of rewritten captions are included in the supplementary material.

% To further ensure visual consistency, we computed CLIP embeddings for images from both SAYCam and our synthetic dataset, selecting pairs through cosine similarity and the sparse Hungarian algorithm. This process produced a dataset closely aligned visually with SAYCam, providing an optimal balance between visual diversity and domain appropriateness. Details of the transformation guidelines and examples of rewritten captions are included in the supplementary material.

\subsection{BabyLLaVA: Generative VLM Baseline}
%Inspired by LLaVA \cite{llava}, our baseline generative model, BabyLLaVA, integrates a language model (GPT-2 with 7.18 million parameters) and a vision encoder (ResNeXt-50 with 23 million parameters) through a lightweight multilayer perceptron (MLP) connector. Training exclusively on our synthesized developmentally plausible dataset, BabyLLaVA serves as the first compact generative model specifically tailored for infant-inspired multimodal tasks. Additional architecture and training details are provided in the supplementary material.
We then train a compact VLM, called BabyLLaVA, using the compiled dataset. Inspired by LLaVA \cite{llava}, BabyLLaVA integrates a compact language model (GPT-2 \cite{radford2019gpt2}, 7M parameters) and vision encoder (ResNeXt-50 \cite{xie2017resnext}, 23M parameters) through a lightweight multilayer perceptron connector. To examine the impact of model capacity, we also provide a larger variant (Llama-1.1B \cite{zhang2024tinyllama} + ViT-L \cite{dosovitskiy2020image}). Consistent with our guiding principles, BabyLLaVA’s compact size and simplified architecture explicitly limit the model's complexity, aligning closely with the realistic developmental constraints of early-stage learners. Training exclusively on our provided developmental dataset, BabyLLaVA provides a suitable baseline model to evaluate the effectiveness of developmentally aligned multimodal learning. Additional details can be found in the supplementary material.

\subsection{Evaluation Tasks}
To rigorously evaluate multimodal reasoning within the intended developmental scope, we introduce three novel benchmarks explicitly designed around cognitive milestones typical of early learners, in addition to the existing Labeled-S task~\cite{orhan2020self}. These tasks deliberately embody simplicity and developmental appropriateness, ensuring alignment with our guiding principles (Figure \ref{fig: benchmark pipeline}).
%To thoroughly evaluate multimodal reasoning capabilities within our developmental domain, we propose three new evaluation benchmarks inspired by infant cognitive milestones, namely Visual Two-Word Test, Baby Winoground, and SAYCam Caption. Alongside Labeled-S \cite{orhan2020self}, these benchmarks constitute our in-domain evaluation suite, as shown in Figure \ref{fig:in-domain benchmarks}.

\myparagraph{Labeled-S.}
The Labeled-S testing dataset was first introduced by Orhan et al. \cite{orhan2020self} and has since been used in studies such as \cite{orhan2024learning, vong2024cvcl}. Leveraging SAYCam annotations, Orhan et al. manually curated a dataset comprising approximately 58K labeled frames across 26 categories. Following Vong et al. \cite{vong2024cvcl} and aligning with standard child testing procedures \cite{mcdaniel1998methods}, we use a subset of Labeled-S and evaluate models by presenting a target category label alongside four candidate images, requiring the model to identify the correct match.

\myparagraph{Visual Two-Word Test (VTWT).} Inspired by the linguistic milestone known as the “two-word” stage (typically 18–24 months) \cite{o2001first, brandone2006language, berk2012two, riccardi2024twowordtest}, this task assesses compositional semantic reasoning. Models must correctly match SAYCam images with appropriate two-word phrases (e.g., “wash cup” vs. “fill cup”). Starting with a sub-sampled test split from SAYCam, we generate 5117 phrase pairs using GPT-4o. These are manually reviewed for linguistic and visual appropriateness, yielding 967 final pairs. Table~\ref{tab:pos_diff} summarizes the distribution of phrase types tested. Detailed annotation guidelines and procedures are provided in the supplementary material.

\begin{table}[htbp]
\centering
\resizebox{0.48\textwidth}{!}{%
\begin{tabular}{lccl}
\toprule
% \textbf{Type of difference} & \textbf{Count} & \textbf{Proportion (\%)} & \textbf{Example} \\
\textbf{Types of Differences} & \textbf{Proportions (\%)} & \textbf{Examples} \\
\midrule
% Verb          & 262            & 27.2                    & ``wash cup'' $\rightarrow$ ``fill cup'' \\
% Adjective     & 206            & 21.4                    & ``happy faces'' $\rightarrow$ ``sad faces'' \\
% Noun          & 164            & 17.0                    & ``car outside'' $\rightarrow$ ``bike outside'' \\
% Verb + Noun   & 206            & 21.4                    & ``spread jam'' $\rightarrow$ ``cut bread'' \\
% Adjective + Noun  & 111        & 11.5                    & ``yellow flower''
%  $\rightarrow$ ``green tree'' \\
% Verb + Adjective  & 14         & 1.5                     & ``small frown'' $\rightarrow$ ``big smile'' \\
Verb       & 27.2                    & ``wash cup'' vs. ``fill cup'' \\
Adjective  & 21.4                    & ``happy faces'' vs. ``sad faces'' \\
Noun       & 17.0                    & ``car outside'' vs. ``bike outside'' \\
Verb + Noun    & 21.4                    & ``spread jam'' vs. ``cut bread'' \\
Adjective + Noun  & 11.5                    & ``yellow flower''
 vs. ``green tree'' \\
Verb + Adjective  & 1.5                     & ``small frown'' vs. ``big smile'' \\
\bottomrule
\end{tabular}%
}
\caption{Proportions and examples of each type of differences between positive and negative phrases in the Visual Two-Word Test.}
\label{tab:pos_diff}
\end{table}

\myparagraph{Baby Winoground.}
Extending VTWT, Baby Winoground tests more advanced visio-linguistic compositional reasoning. Inspired by Winoground \cite{thrush2022winoground}, this task presents two images and two corresponding phrases (one positive, one negative). Negative images were created by modifying specific visual elements of original images through targeted prompts provided to Stability AI’s Stable Image Ultra model \cite{stablediffusion3}. All samples undergo a manual review to ensure minimal domain gaps and precise visio-linguistic mappings, resulting in 365 high-quality test samples. Full prompt-engineering details and validation methods are in the supplementary material.

For evaluation, we adopt a modified group score to better understand model performance under distributional shifts. Each example consists of two image-phrase pairs: one positive pair (a real SAYCam frame and a matching phrase) and one negative pair (a modified phrase and a synthetic image). The standard group score requires the model to correctly identify both the positive and negative pairs over four pairwise comparisons. To gain finer insight, we break this down into two context-conditioned variants:

\begin{itemize}
    \item \textbf{Positive Context Score:} Measures whether the model correctly identifies the matching pair when using the original SAYCam image or phrase as context.
    \item \textbf{Negative Context Score:} Measures the same, but when using the synthetic image or modified phrase as context.
\end{itemize}

\myparagraph{SAYCam Caption.}
The SAYCam Caption benchmark evaluates generative captioning skills by requiring models to generate accurate, contextually relevant descriptions for SAYCam images. Captions are sourced from the test split of SAYCam, using child-directed utterances as ground truth. To refine the dataset, we deduplicate frames with identical utterances, retaining only those with the highest CLIP image-caption similarities. This yields 1598 distinct image-caption pairs, which are then manually verified, resulting in 294 final test samples. Evaluation is performed using the METEOR metric \cite{banerjee2005meteor}. This task measures a model's ability to generate coherent, semantically appropriate child-directed descriptions. Examples of test samples are provided in the supplementary material.

\section{Experiments}

% \begin{table*}[htbp]
%     \centering
%     \resizebox{1.\textwidth}{!}{
%         \begin{tabular}{llcccc}
%         \toprule
%         \textbf{Category} & \textbf{Model} & \textbf{Labeled-S} & \textbf{Visual Two-Word Test} & \textbf{Baby Winoground} & \textbf{SAYCam Caption} \\
%         \midrule
%         \multirow{3}{*}{\textbf{Upper Bound Models}} 
%         & LLaVA-v1.5-7B & 0.7400 & 0.7851 & 0.4274 & 0.1657 \\
%         & LLaVA-v1.5-7B-ft & 0.6591 & 0.7038 & 0.3205 & 0.1798 \\
%         & CLIP-large & 0.7100 & 0.8625 & 0.6740 & N/A \\
%         \midrule
%         \multirow{2}{*}{\textbf{Baby Models}} 
%         & BabyLLaVA & 0.4195 & 0.6252 & 0.0658 & 0.1379 \\
%         & CVCL & 0.6086 & 0.6494 & 0.0932 & N/A \\
%         \midrule
%         \textbf{Random Guess} & - & 0.2500 & 0.5000 & 0.1667 & N/A \\
%         \bottomrule
%         \end{tabular}
%     }
%     \caption{Evaluation results of in-domain tasks, where a higher score indicates better performance. For Labeled-S, we use the same target and foil testing samples as \cite{orhan2020self} and report accuracy. For the Visual Two-Word Test, we report accuracy. For Baby Winoground, we report the group score only. For SAYCam Caption, we report the METEOR score.}
%     \label{tab:main_result_part1}
% \end{table*}

\begin{table*}[htbp]
    \centering
    \begin{tabularx}{1.\textwidth}{llccXXXc}
        \toprule
        \textbf{Category} & \textbf{Model} & \textbf{Labeled-S} & \textbf{VTWT} & \multicolumn{3}{c}{\textbf{Baby Winoground}} & \makecell{\textbf{SAYCam}\\\textbf{Caption}} \\
        \cmidrule(lr){5-7}
        & & & & Overall & Pos. Ctx & Neg. Ctx & \\
        \midrule
        \multirow{3}{*}{\makecell{\textbf{Upper Bound} \\ \textbf{Models}}}
        & LLaVA-v1.5-7B & 0.7400 & 0.7851 & 0.4274 & 0.6575 & 0.6301 & 0.1657 \\
        & LLaVA-v1.5-7B-ft & 0.6591 & 0.7038 & 0.3205 & 0.5644 & 0.6027 & 0.1798 \\
        & CLIP-large & 0.7100 & 0.8625 & 0.6740 & 0.7315 & 0.8603 & N/A \\
        \midrule
        \multirow{3}{*}{\textbf{Baby Models}}
        & BabyLLaVA-GPT2 & 0.4195 & 0.6252 & 0.0658 & 0.3890 & 0.2301 & \textbf{0.1379}\\
        & BabyLLaVA-Llama & 0.4200 & 0.6029 & 0.0521 & 0.3945 & \textbf{0.2466} & 0.1287\\
        & CVCL & \textbf{0.6086} & \textbf{0.6494} & \textbf{0.0932} & \textbf{0.5068} & 0.2246 & N/A \\
        \midrule
        \textbf{Random Guess} & - & 0.25 & 0.5 & 0.1667 & 0.25 & 0.25 & N/A \\
        \bottomrule
    \end{tabularx}
    \caption{Evaluation results of in-domain tasks, where a higher score indicates better performance. For Labeled-S, we use the same target and foil testing samples as \cite{orhan2020self} and report accuracy. For the Visual Two-Word Test (VTWT), we report accuracy. For Baby Winoground, we report the group score for different contexts. For SAYCam Caption, we report the METEOR score.}
    \label{tab:main_result_part1}
\end{table*}

Our experimental evaluation aims to compare VLM architectures and training paradigms within a developmentally plausible setting, investigate the effectiveness of our synthetic child-directed dataset, and perform a fine-grained analysis of compositional reasoning. To validate that our baby models align with the cognitive and linguistic limitations of early-stage learners, we also assess baby models on tasks that exceed typical infant-level developmental capacities, presented in the supplementary materials.

\subsection{In-Domain Benchmark Results}
\label{sec:in_domain_results}
We begin by evaluating multiple models, including baby models trained purely on SAYCam (BabyLLaVA, CVCL~\cite{vong2024cvcl}) and larger upper bound models that are either directly used out of the box (LLaVA-v1.5-7B~\cite{llava1.5}, CLIP-large~\cite{CLIP}) or further fine-tuned on our SAYCam data (LLaVA-v1.5-7B-ft). These models are assessed on four in-domain benchmarks: Labeled-S, Visual Two-Word Test (VTWT), Baby Winoground, and SAYCam Caption. \mbox{Table}~\ref{tab:main_result_part1} summarizes these results.

Notably, CVCL—a contrastive model—consistently outperforms the generative BabyLLaVA model across most tasks. This observation aligns with existing literature~\cite{zhang2024why, teterwak2023clamp, he2025analyzing, geigle2024african}, suggesting that contrastive models may be better suited to discriminative tasks, possibly due to their direct objective of learning joint visual-textual alignment. However, generative models like BabyLLaVA demonstrate reasonable performance on simpler compositional tasks such as VTWT, indicating substantial potential for improvement on more sophisticated compositional tasks like Baby Winoground. Interestingly, the larger \mbox{BabyLLaVA-Llama} variant performs similarly or even worse than \mbox{BabyLLaVA-GPT2}, despite being about 50 times larger—suggesting overfitting to the limited data; as such, we default to using the GPT2 version for all the remaining discussion. In particular, Baby Winoground reveals a stark asymmetry: baby models perform above chance when reasoning from in-distribution (positive) context, but below chance from out-of-distribution (negative) context, highlighting a systematic failure under distribution shift. Moreover, generative captioning, measured by \mbox{SAYCam} Caption scores, remains challenging for all models, emphasizing the additional complexity inherent in generating full linguistic descriptions from minimal data.
% \begin{table*}[htbp]
%     \centering
%     \resizebox{1.\textwidth}{!}{
%         \begin{tabular}{lcccc}
%         \toprule
%         \textbf{Model} & \textbf{Labeled-S} & \textbf{Visual Two-Word Test} & \textbf{Baby Winoground} & \textbf{SAYCam Caption} \\
%         \midrule
%         CVCL-ori & 0.6086 & 0.6494 & 0.0932 & N/A \\
%         CVCL-aug & 0.5805 & 0.7021 & 0.2027 & N/A \\
%         CVCL-aug-random & 0.6023 & 0.6835 & 0.1068 & N/A \\
%         \midrule
%         BabyLLaVA-ori & 0.4195 & 0.6252 & 0.0658 & 0.1379 \\
%         BabyLLaVA-aug & 0.5364 & 0.6933 & 0.0822 & 0.1592 \\
%         BabyLLaVA-aug-random & 0.5155 & 0.6553 & 0.0877 & 0.1778 \\
%         \bottomrule
%         \end{tabular}
%     }
%     \caption{Ablation study of our transferred dataset on in-domain tasks. \textbf{XX-ori:} Only SAYCam. \textbf{XX-aug:} SAYCam + child-directed data. \textbf{XX-aug-random:} SAYCam + random general purpose data.}
%     \label{tab:dataset abalation}
% \end{table*}
% \subsection{Effectiveness of the Transferred Dataset}
% \label{sec:dataset_ablation}

\begin{table*}[htbp]
    \centering
    \begin{tabularx}{\textwidth}{lccXXXc}
        \toprule
        \textbf{Model} & \textbf{Labeled-S} & \textbf{VTWT} & \multicolumn{3}{c}{\textbf{Baby Winoground}} & \makecell{\textbf{SAYCam}\\\textbf{Caption}} \\
        \cmidrule(lr){4-6}
        & & & Overall & Pos. Ctx & Neg. Ctx & \\
        \midrule
        % \multirow{3}{*}{\makecell{\textbf{Upper Bound} \\ \textbf{Models}}}
        CVCL-filtered & 0.6086 & 0.6494 & 0.0932 & 0.5068 & 0.2246 & N/A \\
        CVCL-filtered-aug & 0.5805 & 0.7021 & 0.2027 & 0.4657 & 0.4493 & N/A \\
        CVCL-filtered-random & 0.6023 & 0.6835 & 0.1068 & 0.4739 & 0.2958 & N/A \\
        \midrule
        % \multirow{2}{*}{\textbf{Baby Models}}
        BabyLLaVA-filtered & 0.4195 & 0.6252 & 0.0658 & 0.3890 & 0.2301 & 0.1379 \\
        BabyLLaVA-filtered-aug & 0.5364 & 0.6933 & 0.0822 & 0.3726 & 0.3096 & 0.1592 \\
        BabyLLaVA-filtered-random & 0.5155 & 0.6553 & 0.0877 & 0.3616 & 0.2712 & 0.1778 \\
        BabyLLaVA-filtered-double & 0.4659 & 0.6383 & 0.0658 & 0.2411 & 0.2301 & 0.1799 \\
        BabyLLaVA-aug-only & 0.5000 & 0.6239 & 0.0630 & 0.3370 & 0.3644 & 0.0615 \\
        BabyLLaVA-random-only & 0.4400 & 0.5098 & 0.0548 & 0.2904 & 0.3836 & 0.0615 \\
        % \midrule
        % \textbf{Random Guess} & - & 0.25 & 0.5 & 0.1667 & 0.25 & 0.25 & N/A \\
        \bottomrule
    \end{tabularx}
    \caption{Ablation study of our transferred dataset on in-domain tasks. \textbf{XX-filtered:} Only SAYCam. \textbf{XX-filtered-aug:} SAYCam + child-directed transferred data. \textbf{XX-filtered-random:} SAYCam + random general-domain data. \textbf{XX-filtered-double:} SAYCam with double-numbered samples. \textbf{XX-aug-only:} Only random general-domain data. \textbf{XX-random-only:} Only random general-domain data.}
    \label{tab:dataset abalation}
\end{table*}

\subsection{Transferred Dataset Ablation}

We next perform an ablation study (Table~\ref{tab:dataset abalation}) comparing models trained on several different dataset settings: using our filtered SAYCam dataset only (\textit{filtered-only}), SAYCam plus our transferred, child-directed dataset (\textit{filtered-aug}), transferred dataset only (\textit{aug-only}), SAYCam plus randomly selected general-domain dataset of the same size as the transferred dataset (\textit{filtrered-random}), random general-domain dataset only (\textit{random-only}), and a bigger filtered SAYCam dataset whose number of samples is doubled by relaxing the threshold (\textit{filtered-double}). Additional analysis regarding data efficiency can be found in the supplementary material.

We observe clear performance improvements in CVCL and BabyLLaVA when using our carefully curated dataset compared to random augmentation, particularly on compositional reasoning tasks such as VTWT and Baby Winoground. These results indicate that explicitly adapting general-domain data to reflect the linguistic simplicity and visual content of infant environments significantly enhances the data efficiency and overall alignment of the resulting models. Using only the transferred data degrades the model performance, which shows the importance of having the original SAYCam dataset as an anchor. Notably, for Baby Winoground, training with the transferred dataset substantially improves performance in the negative context setting, despite a slight drop in the positive context score; while the randomly selected dataset also improves the negative context score, the gains are smaller, indicating that our transferred dataset is more effective in helping baby models generalize to broader domains. In contrast, improvements in generative captioning remain modest, suggesting that further refinements, such as enriching the linguistic variety or introducing narrative structures, could improve generative performance.

\subsection{Assessing Language Bias in VTWT}
\label{sec:vtwt_language_bias}
To confirm the robustness of our VTWT benchmark, we conducted an experiment removing visual context entirely (Table~\ref{tab:language_prior_ablation}). The resulting performance drop from around 78\% accuracy (with image) to approximately random chance (53\% without image) demonstrates that the task cannot be solved through language biases alone. This validates VTWT as a rigorous evaluation of genuine multimodal compositional reasoning rather than simple linguistic pattern-matching, confirming the robustness and appropriateness of our benchmark.
\begin{table}[htbp]
    \centering
    \resizebox{0.5\textwidth}{!}{
        \begin{tabular}{lcc}
        \toprule
        \textbf{Model} & \textbf{VTWT (w/ image)} & \textbf{VTWT (w/o image)} \\
        \midrule
        LLaVA-v1.5-7B & 0.7851 & 0.5307 \\
        BabyLLaVA & 0.6252 & 0.5360 \\
        \bottomrule
        \end{tabular}
    }
    \caption{Ablation study of language-only bias on VTWT.}
    \label{tab:language_prior_ablation}
\end{table}
%Interestingly, we observe a much smaller gap between BabyLLaVA and the larger LLaVA-v1.5-7B model on the language-only setting (53.60\% vs. 53.07\%), compared to the multimodal condition (62.52\% vs. 78.51\%). This smaller gap indicates that differences in performance are not solely attributable to model capacity but also significantly influenced by the complexity of the multimodal task and the specific nature of the developmental data. Consequently, attributing performance gaps purely to model size or capacity would overlook critical factors such as data alignment and task complexity.

\subsection{Investigating Compositional Reasoning}
\label{sec:compositional_analysis}
We further dissect the VTWT performance by examining model accuracy on different types of compositional differences (noun, verb, adjective, or combinations thereof) in \mbox{Table}~\ref{tab:model_perf_twt-fine}.
\begin{table}[h]
\centering
\resizebox{0.48\textwidth}{!}{%
\begin{tabular}{lccc}
\toprule
\textbf{Type of Difference} & \textbf{CVCL-filtered} & \textbf{CVCL-filtered-aug} & \textbf{CVCL-filtered-random} \\
\midrule
Verb              & 0.6221 & 0.6564 & 0.7404 \\
Adjective         & 0.5507 & 0.6086 & 0.5797 \\
Noun              & 0.7317 & 0.7682 & 0.7073 \\
Verb + Noun       & 0.7087 & 0.7572 & 0.6699 \\
Adjective + Noun  & 0.6936 & 0.7927 & 0.7297 \\
Verb + Adjective  & 0.4285 & 0.6428 & 0.7857 \\
\bottomrule
\end{tabular}%
}

\caption{Performance breakdown on VTWT by part-of-speech differences.}
\label{tab:model_perf_twt-fine}
\end{table}

% \begin{table*}[htbp]
%     \centering
%     \resizebox{\textwidth}{!}{%
%         \begin{tabular}{llccccc}
%         \toprule
%         \textbf{Category} & \textbf{Model} & \textbf{BLiMP\textsubscript{filtered}} & \textbf{BLiMP\textsubscript{supplement}} & \textbf{Winoground} & \textbf{VQA} & \textbf{DevBench} \\
%         \midrule
%         \multirow{3}{*}{\textbf{Upper Bound Models}} 
%         & LLaVA-v1.5-7B & 0.7299 & 0.8300 & 0.6327 & 0.6273 & 0.8570 \\
%         & LLaVA-v1.5-7B-ft & 0.7205 & 0.8032 & 0.5992 & 0.4941 & 0.6300 \\
%         & CLIP-large & N/A & N/A & 0.5638 & 0.2397 & 0.7172 \\
%         \midrule
%         \multirow{2}{*}{\textbf{Baby Models}} 
%         & BabyLLaVA & 0.6772 & 0.5903 & 0.5214 & 0.2312 & 0.3907 \\
%         & CVCL & N/A & N/A & 0.5221 & 0.1600 & 0.3993 \\
%         \midrule
%         \textbf{Random Guess} & - & 0.5000 & 0.5000 & 0.5000 & 0.1250 & 0.3750 \\
%         \bottomrule
%         \end{tabular}%
%     }
%     \caption{Evaluation results on out-of-domain benchmarks. For BLiMP, Winoground and VQA, please refer to \cite{choshen2024babylmchallenge} for implementation details. For DevBench, we report the average score of TROG, WG, LWL and VV.}
%     \label{tab:main_result_part2}
% \end{table*}

We observe that all three model variants perform worse on adjective differences than adjective + noun differences. We suspect this is because adjective differences alone are often less visually explicit in an image, and the presence of an additional noun difference helps the models disambiguate these cases.

Additionally, models trained exclusively on developmentally plausible data (CVCL-filtered and CVCL-filtered-aug) exhibit distinct performance patterns. For single-component differences (the first three rows of Table \ref{tab:model_perf_twt-fine}), both models achieve their highest performance on noun differences and perform worse on verb and adjective differences (e.g., 76\% vs. 65\% and 60\% respectively for CVCL-filtered-aug). This result aligns with linguistic development findings from \cite{berk2012two, sandhofer2007learning}, which suggest that early-stage language learners use nouns at least twice as often as verbs and are also slower to acquire adjectives. However, similar phenomenon is not observed from CVCL-filtered-random.

The alignment between our empirical results and developmental psychology literature reinforces that our targeted synthetic data transformations effectively facilitate more robust baby-like representations — a central objective identified in our introduction.

\subsection{Discussion}
Overall, our experiments reinforce several key insights central to our initial narrative. First, child-directed transformations of general-domain datasets provide substantial gains within our developmentally plausible domain, validating our approach. However, generative models face heightened difficulties in compositional reasoning and full-sentence generative tasks, highlighting significant room for further development. Lastly, while the specialized infant-oriented approach offers promising efficiencies, it inherently limits performance in broader contexts. These findings suggest fruitful future directions, including expanding dataset richness, exploring hybrid generative-discriminative training methods, and generalizing our approach to other specialized domains, as envisioned in our introduction.

\section{Conclusion and Future Work}
\label{sec:conclusion}

In this work, we introduced \ours{}, a framework for data-efficient pretraining and evaluation of compact vision-language models (VLMs) inspired by the developmental learning conditions of human infants. Our approach is grounded in explicitly enforcing developmental constraints on both data and model design, ensuring that baby models operate within a realistic cognitive scope. To achieve this, we curated a filtered subset of the SAYCam dataset, constructed a novel synthetic training dataset that aligns with child-directed language and visual experiences, and introduced three evaluation benchmarks designed to test multimodal reasoning at early developmental stages.

Our experiments validate the effectiveness of this approach. In-domain evaluations demonstrate that baby models can learn meaningful multimodal associations from developmentally appropriate data, while out-of-domain evaluations confirm their intentional constraints, preventing overgeneralization beyond early cognitive capabilities. Notably, we find that the observed performance gaps between baby models and larger models arise from multiple factors—model capacity, task complexity, and data alignment—rather than capacity alone. This underscores the importance of dataset and task design in modeling early-stage learning.

Moving forward, our work opens several avenues for further research. First, expanding the dataset to incorporate additional multimodal learning signals—such as temporal context or richer object interactions—could further refine developmental modeling. Second, investigating hybrid models that balance generative and contrastive training may provide insights into optimizing learning efficiency in data-limited regimes. Lastly, our benchmarks and methodology can serve as a foundation for broader inquiries into how developmental constraints shape representation learning in neural models.

By establishing a principled framework for modeling early-stage multimodal learning, \ours{} provides a meaningful step toward understanding and replicating data-efficient learning in artificial systems, with potential implications for both machine learning and cognitive science.

{
    \small
    \bibliographystyle{ieeenat_fullname}
    \bibliography{main}
}

\clearpage
\onecolumn
\setcounter{page}{1}
\maketitlesupplementary
\normalsize  % Ensure normal font size is restored

\raggedright

% Set the table counter to start from 7
\setcounter{table}{6}  % 1 less than the initial value

% Set the figure counter to start from 3
\setcounter{figure}{3} % 1 less than the initial value

\paragraph{Examples of Filtered SAYCam Dataset.}
The filtered SAYCam training dataset consists of 67,280 image-utterance pairs in total. We provide some examples below.

\begin{figure}[htbp]
    \centering
    \includegraphics[width=0.93\linewidth]{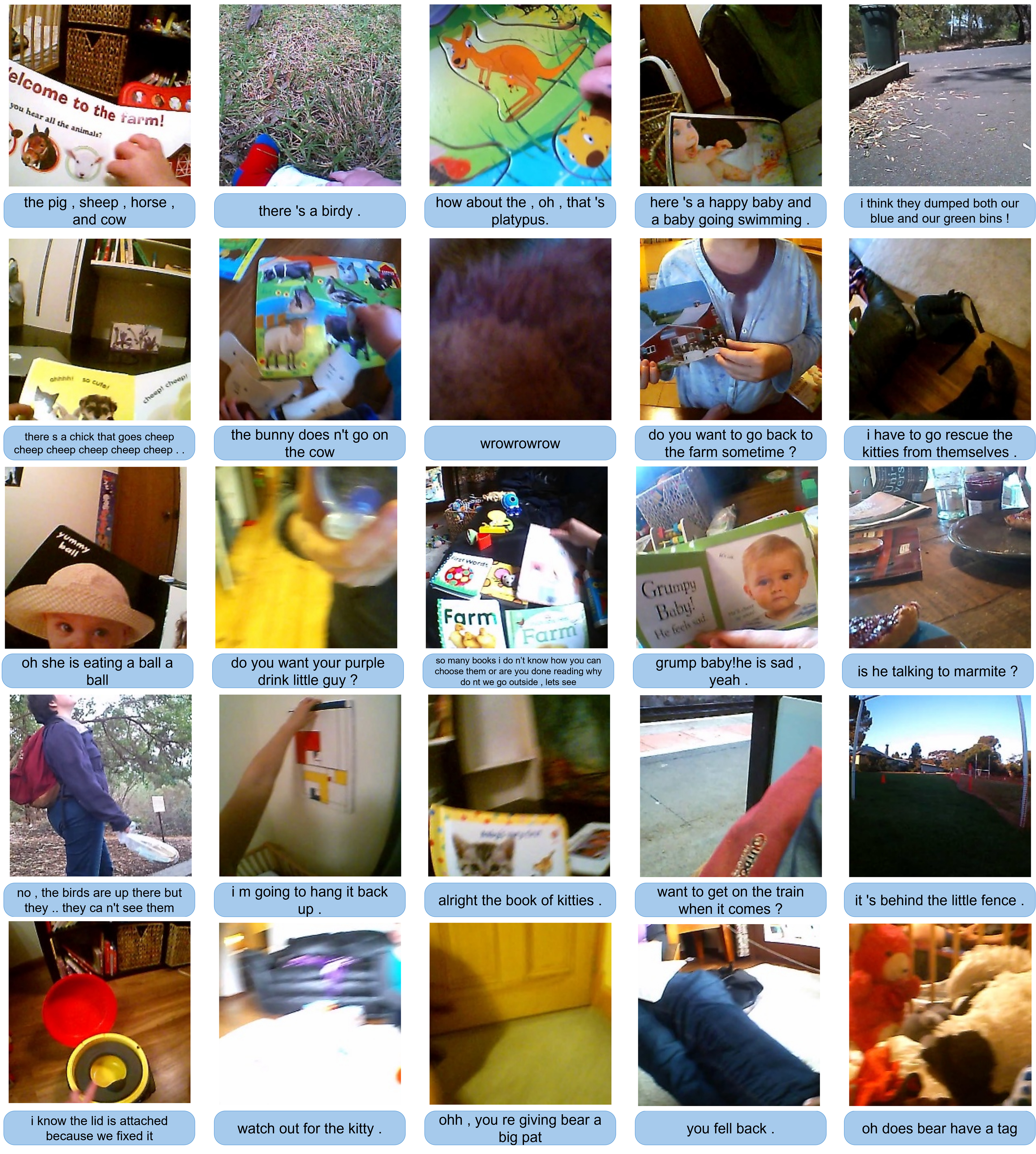}
    \caption{Examples of the filtered SAYCam dataset}
    \label{fig:SAYCam examples}
\end{figure}

\clearpage
\paragraph{Implementation Details for Creating Transferred Training Dataset.}

Starting from LLaVA’s pretraining dataset \cite{llava}, which includes approximate 558K image-caption pairs coming from CC3M \cite{CC3M}, LAION \cite{schuhmann2022laion}, and SBU \cite{ordonez2011im2text}, we carefully design few-shot examples to prompt GPT-4o to transform original captions into simple, natural utterances that a caregiver might say to a two-year-old. Additionally, we instruct GPT-4o to identify captions misaligned with a child's daily experience by explicitly outputting an infeasibility flag in its JSON mode. We get 339,826 feasible samples after this step. The detailed prompt for GPT-4o is provided below.

\begin{figure*}[htbp]
    \centering
    \includegraphics[width=1.\linewidth]{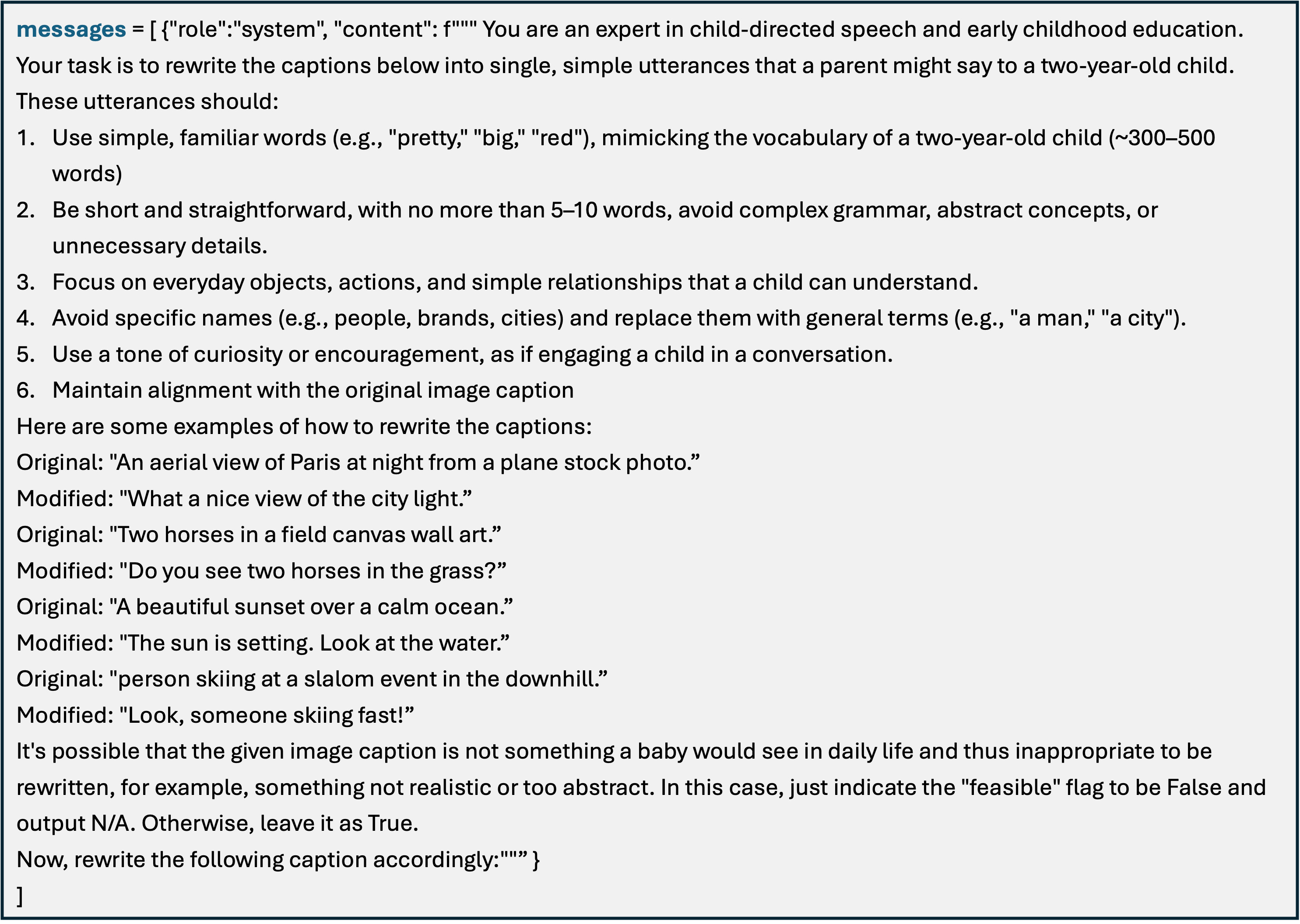}
    \caption{Full prompt for transferred dataset creation}
    \label{fig:td prompt}
\end{figure*}

\begin{figure}[htbp]
    \centering
    \includegraphics[width=1\linewidth]{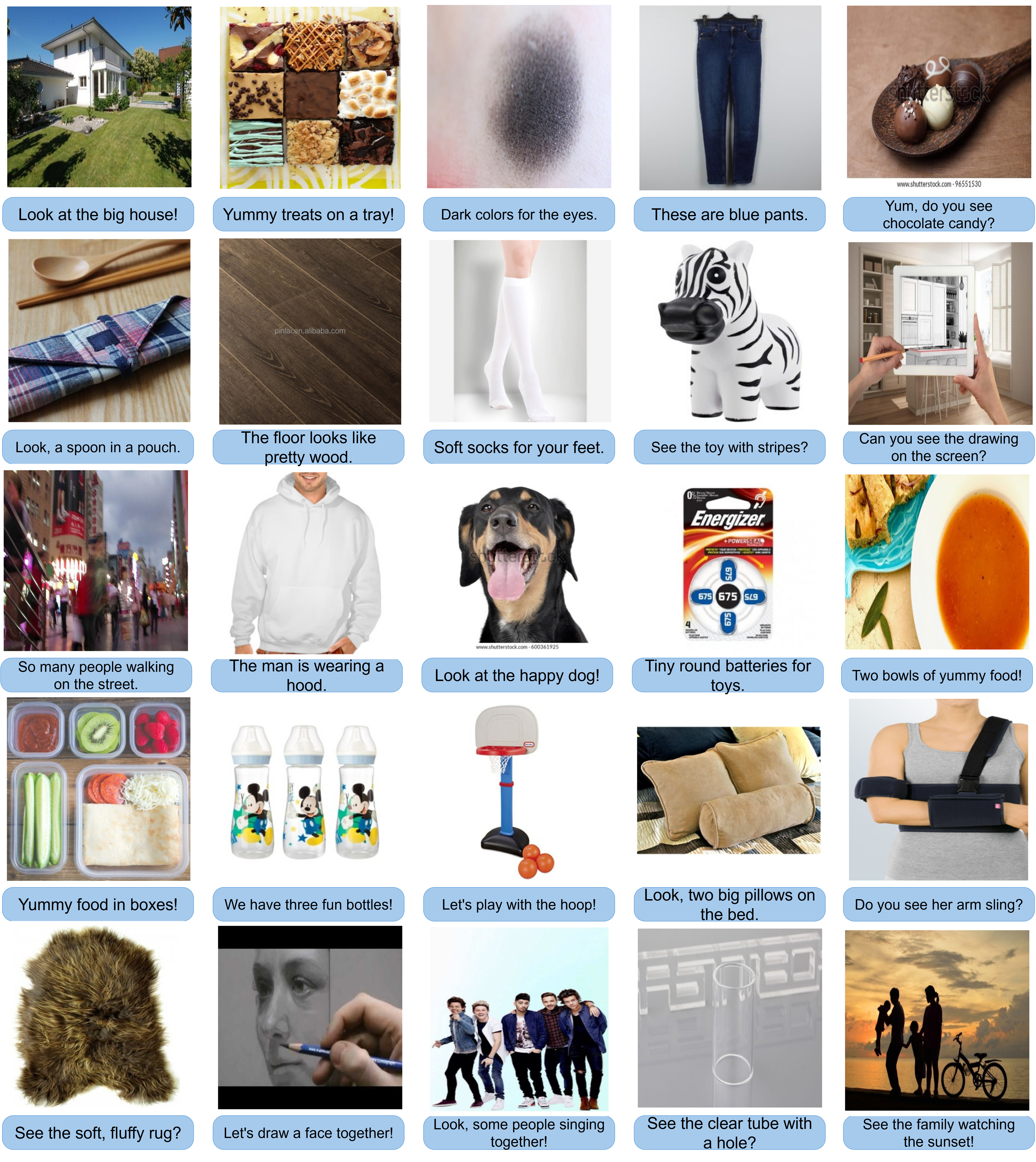}
    \caption{Examples of the transferred dataset}
    \label{fig:transferred dataset example}
\end{figure}

To enhance visual consistency, we use CLIP similarity \cite{CLIP} to select a subset of samples matching the size of the filtered SAYCam training dataset. Specifically, we compute CLIP similarity between each image in the filtered SAYCam dataset and every image in the transferred LLaVA pretraining dataset. Given the significantly larger size of the latter, we retain only the top 1,000 most similar images for each SAYCam image, setting the similarity of all others to zero, resulting in a sparse similarity matrix. We then apply the sparse Hungarian algorithm \cite{jonker1988shortest} to establish a one-to-one match between images from the transferred dataset and the filtered SAYCam. Examples of the final transferred dataset can be seen in Figure \ref{fig:transferred dataset example} on the next page.

\begin{figure*}[htbp]
    \centering
    \includegraphics[width=1\linewidth]{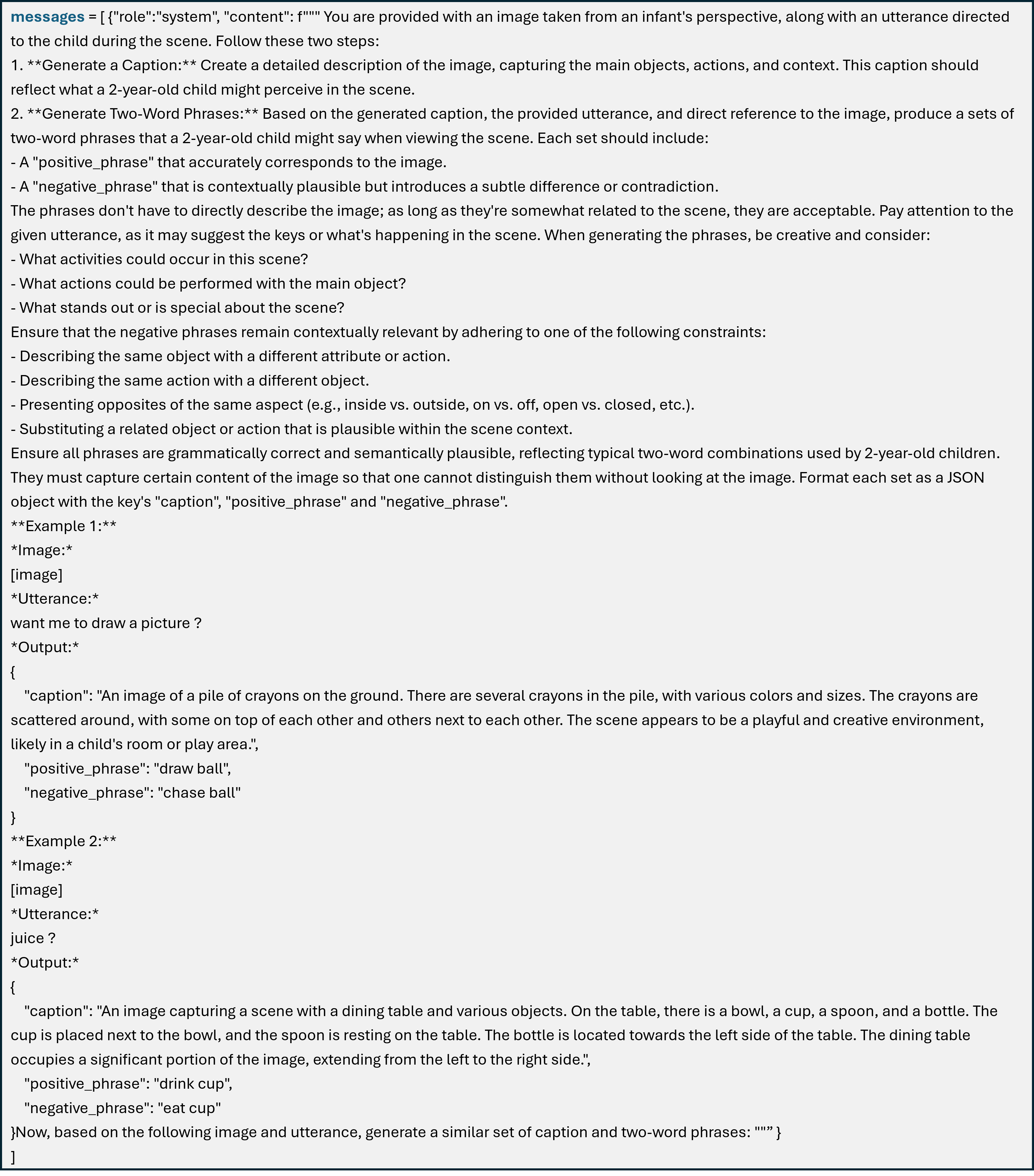}
    \caption{Full prompt for VTWT}
    \label{fig:vtwt prompt}
\end{figure*}

\clearpage
\paragraph{Implementation Details for Creating the Visual Two-Word Test.}
We construct VTWT by sub-sampling the SAYCam test split and using GPT-4o to generate 5,117 candidate two-word phrases through structured prompts. These prompts incorporate few-shot examples and Chain-of-Thought (CoT) guidance to enhance phrase quality. Specifically, we first prompt GPT-4o to generate a detailed image description based on the input image and utterance. Using this description, the model then generates a pair of two-word phrases—one positive and one negative—that differ in noun, verb, or adjective, ensuring clear semantic distinctions. The detailed prompt, along with few-shot examples, is shown in Figure \ref{fig:vtwt prompt}.

To ensure the quality of the test samples, each sample was manually reviewed by two expert annotators with experience in vision and language research to verify that: (1) the caption is correctly describing the image in detail, (2) the positive phrase is concretely depicted in the image, (3) the negative phrase is not depicted in the image, and (4) both the positive and negative phrases are linguistically plausible. After this review, 967 high-quality test samples remain in the benchmark. Examples of VTWT test samples are shown in Figure \ref{fig:vtwt examples} below.

\begin{figure*}[htbp]
    \centering
    \includegraphics[width=1\linewidth]{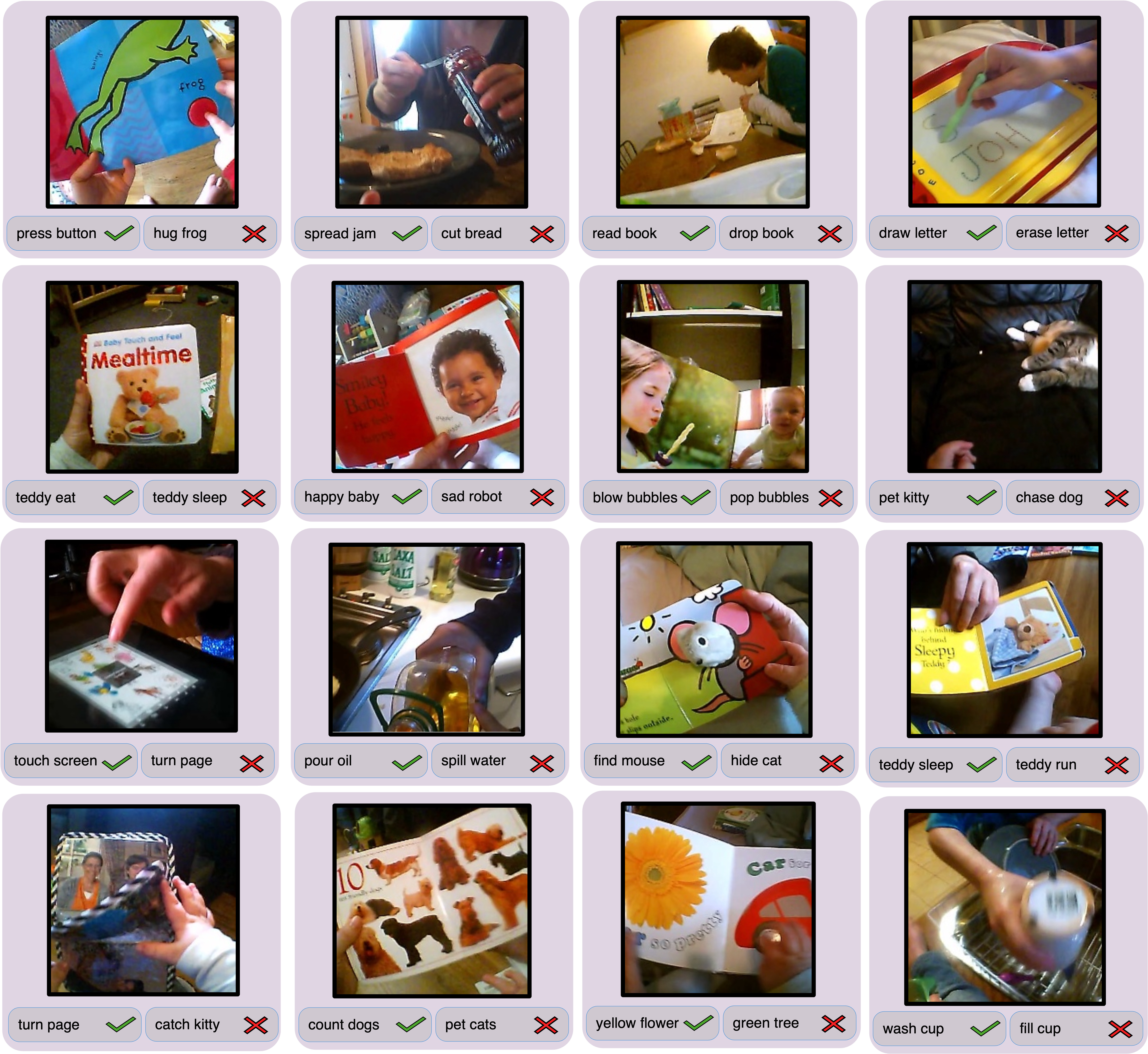}
    \caption{Examples of VTWT Task}
    \label{fig:vtwt examples}
\end{figure*}

\begin{figure*}[htbp]
    \centering
    \includegraphics[width=1\linewidth]{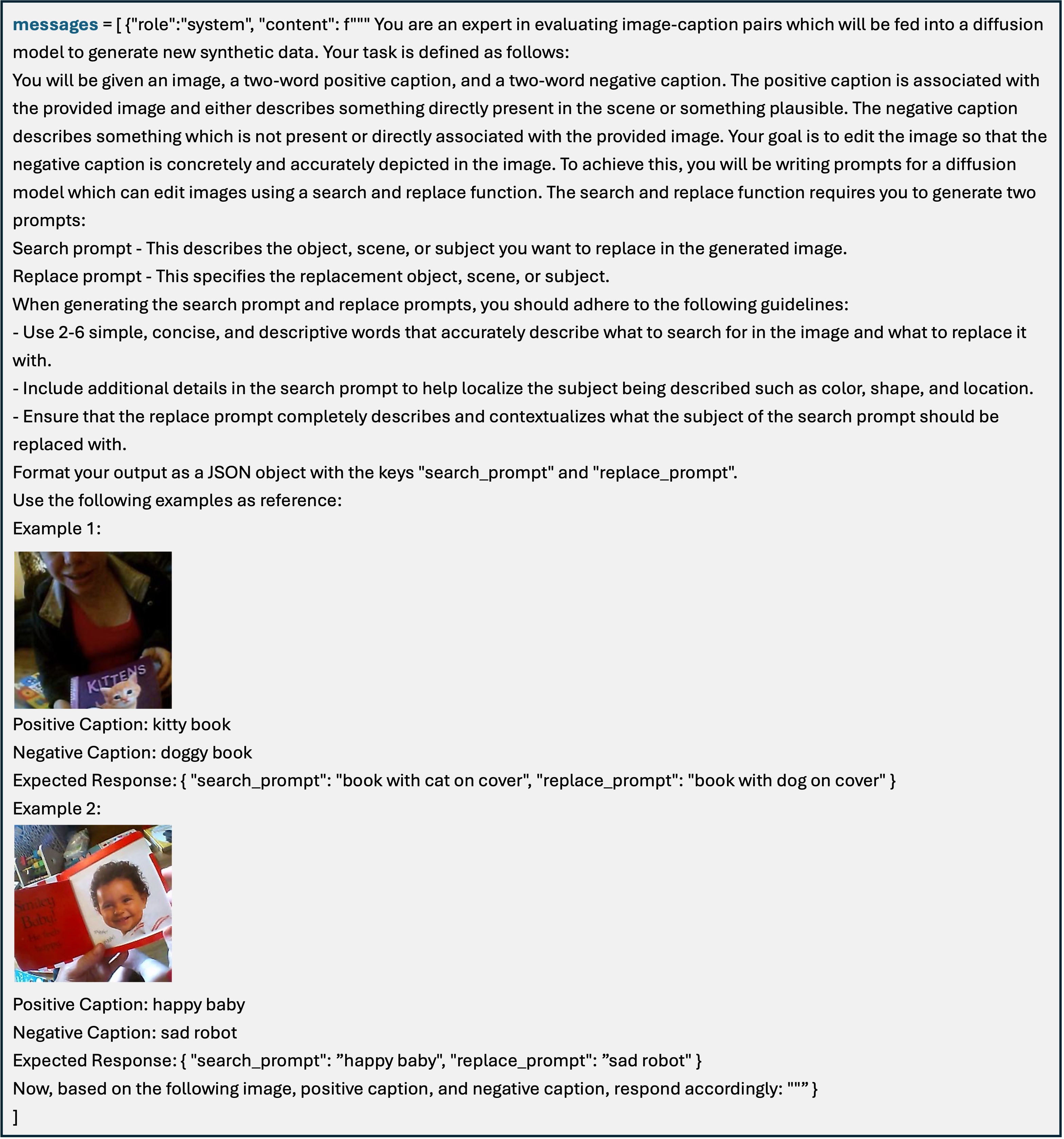}
    \caption{Full prompt for Baby Winoground. We uses few-shot examples to generate search and replace prompts for the image-editing model. }
    \label{fig:bw prompt}
\end{figure*}

\begin{figure}[htbp]
    \centering
    \includegraphics[width=0.85\linewidth]{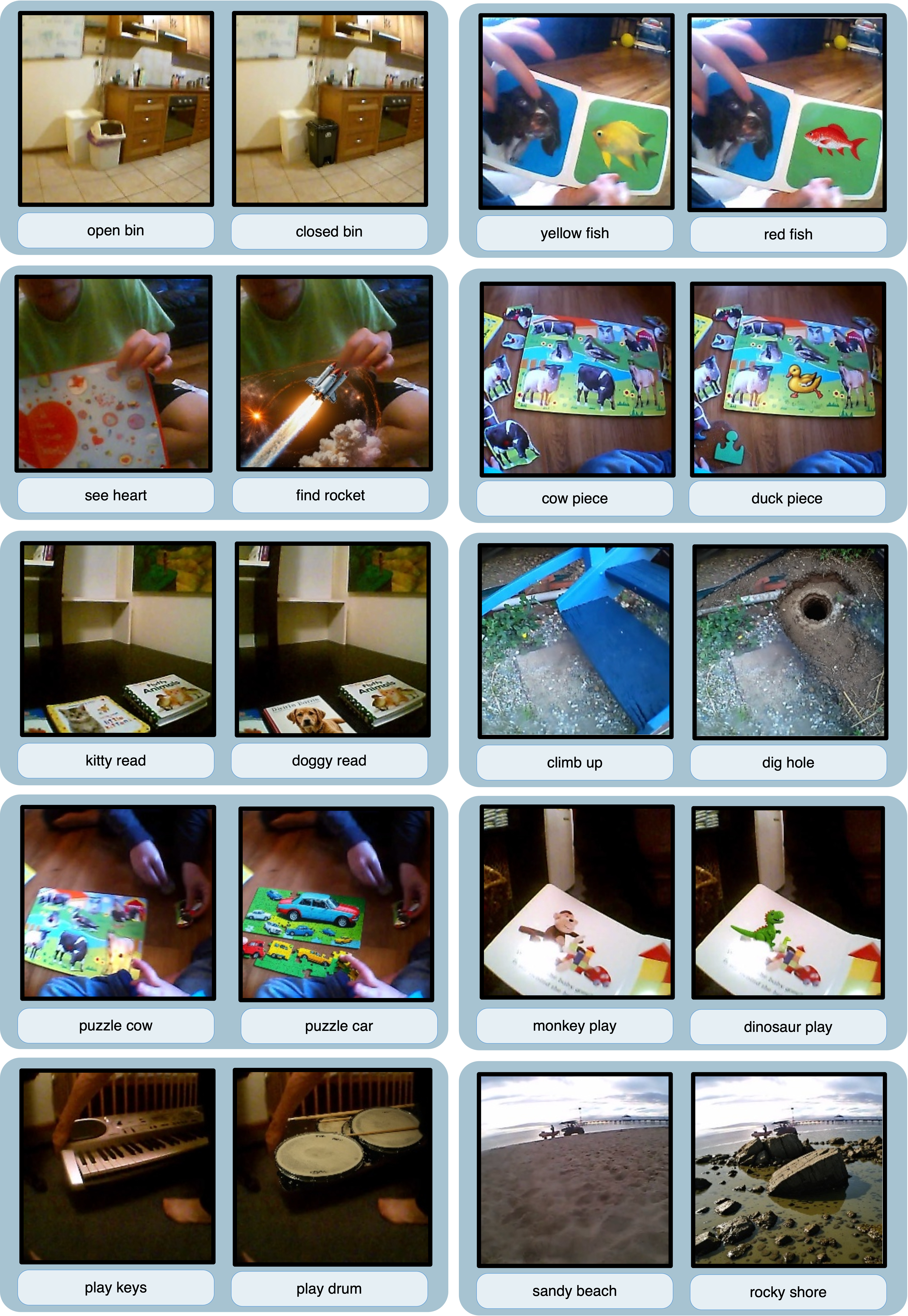}
    \caption{Examples of Baby Winoground Task}
    \label{fig:bw examples}
\end{figure}

\clearpage
\paragraph{Implementation Details for Creating Baby Winoground.} We construct Baby Winoground using the 967 test samples from the Visual Two-Word Test (VTWT). Our goal is to modify the original image from VTWT such that the modified image is associated exclusively with the negative phrase while preserving most of the original content. To achieve this, we leverage the \textit{search and replace} functionality of Stability AI's Stable Image Ultra model~\cite{stablediffusion3} as our image-editing tool. This process requires two prompts:
\begin{itemize}
    \item Search Prompt: Describes the object, subject, or scene to be replaced in the image.
    \item Replace Prompt: Specifies the new object, subject, or scene replacing the original.
\end{itemize}
A direct approach would be to use the positive and negative phrases as search and replace prompts, respectively. However, the two-word constraint often omits crucial details, making it difficult for the image-editing model to generate accurate edits. To address this, we prompt GPT-4o to dynamically generate more descriptive search and replace prompts. We provide few-shot examples and specify key characteristics empirically found to improve edit quality; the full prompt is shown in Figure \ref{fig:bw prompt}. As in VTWT, expert annotators manually review all test samples, ensuring that the edited images align exclusively with the negative phrases. After filtering, 365 high-quality test samples remain. Examples are shown in Figure \ref{fig:bw examples}.

\clearpage
\paragraph{Examples of SAYCam Caption.} The SAYCam Caption task consists of 294 test samples in total. We provide some examples below.

\begin{figure}[htbp]
    \centering
    \includegraphics[width=0.93\linewidth]{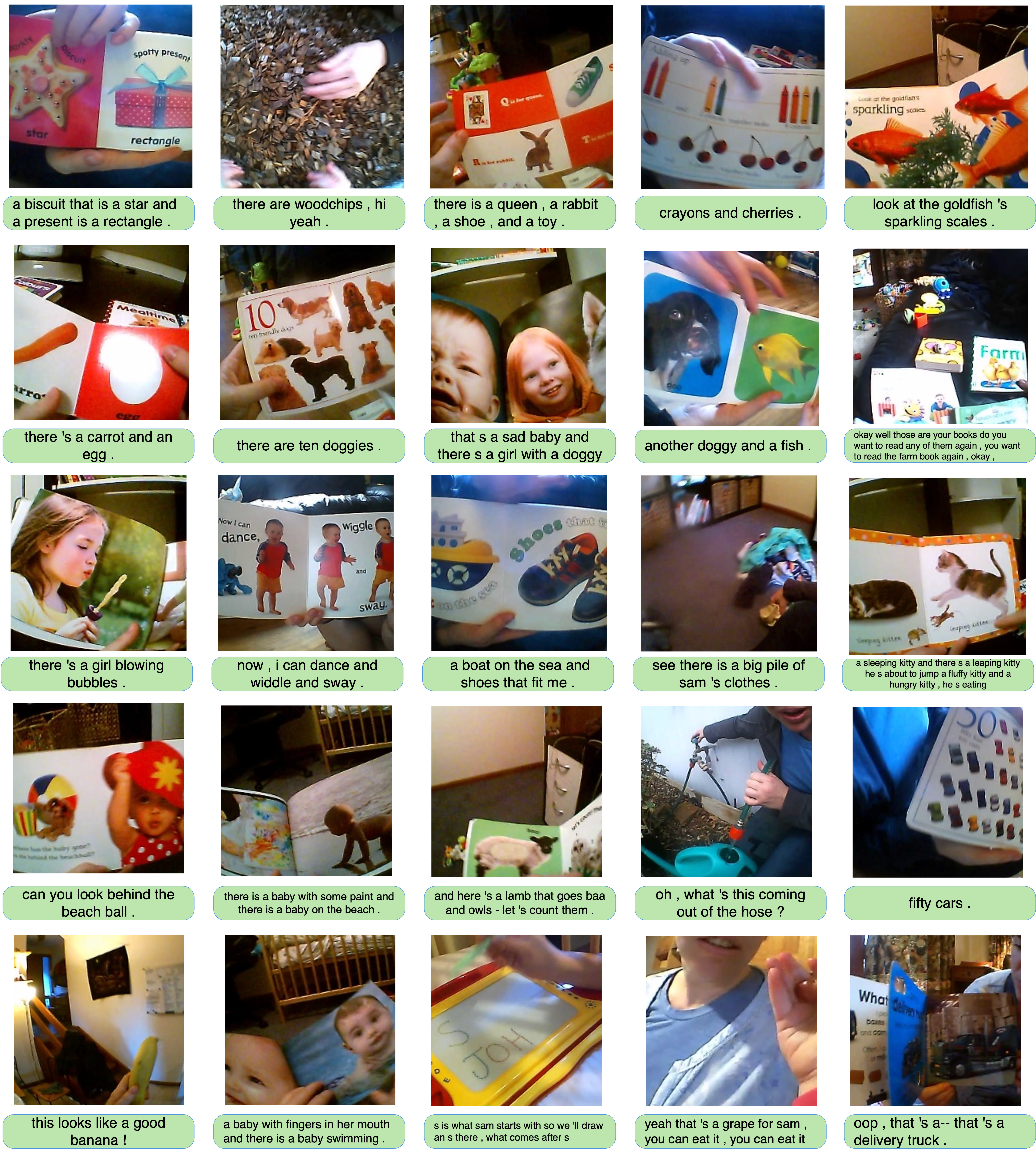}
    \caption{Examples of SAYCam Caption task}
    \label{fig:SAYCam examples}
\end{figure}

\clearpage
\paragraph{BabyLLaVA Training and Evaluation Details.} BabyLLaVA follows the architecture and training strategy of LLaVA~\cite{llava, llava1.5}, consisting of a language backbone, a vision backbone, and a two-layer MLP connector.

For the language backbone, we train a small GPT-2 model with 7.18M parameters from scratch using the language portion of our training corpus. The vision backbone is directly adopted from Orhan et al. \cite{orhan2024learning} and is based on a ResNeXt-50 \cite{xie2017aggregated} model with 23 million parameters, trained from scratch using DINOv2 \cite{oquab2023dinov2} on all SAYCam video clips, including those without utterance transcriptions. These clips are subsampled into 9 million frames at 5 FPS. The connector is a simple two-layer MLP, identical to LLaVA-v1.5.

Our training framework closely follows LLaVA but introduces Stage 0, an additional unimodal pretraining stage for the language and vision backbones. Unlike LLaVA, which initializes from pretrained CLIP and Vicuna v1.5 \cite{zheng2023judging}, BabyLLaVA requires this extra stage since both backbones are trained from scratch. The full training process consists of the following three stages. All the stages can be finished within 2 hours on four A6000 GPUs.
\begin{itemize}
    \item \textbf{Stage 0: Unimodal Pretraining.} The language backbone is trained independently on textual data, while the vision backbone remains unchanged, as we adopt the pretrained backbone directly from Orhan et al.
    \item \textbf{Stage 1: Feature Alignment.} Both backbones are frozen, and only the MLP connector is trained to align vision and language features.
    \item \textbf{Stage 2: End-to-End Training.} The vision backbone remains frozen while the connector and language backbone are trained jointly. We also experiment with different freezing strategies (freezing only the vision backbone, only the language backbone, or neither) and find that freezing only the vision backbone yields the best overall performance.
\end{itemize}

For evaluation, we observe that the choice of input prompt significantly impacts performance, a phenomenon noted in prior research \cite{liu2022design, brown2020language}. To investigate this effect, we test various prompts, including common patterns of child-directed utterances (e.g., ``Look at'' or ``What's that''), as well as the absence of a prompt. Interestingly, omitting the input prompt yields the best results, likely because it aligns with the model's training setup, which does not incorporate fixed prompts.

% We conduct a hyperparameter sweep to determine the optimal training configuration for the filtered SAYCam dataset. The best-performing setup is summarized in Table~\ref{tab:hyperparams}.

% \begin{table}[h]
%     \centering
%     \label{tab:hyperparams}
%     \begin{tabular}{l|ccc}
%         \toprule
%         \textbf{Hyperparameter} & \textbf{Stage 0} & \textbf{Stage 1} & \textbf{Stage 2} \\
%         \midrule
%         Batch size & 256 & 256 & 128 \\
%         Learning rate (lr) & 1e-3 & 1e-3 & 2e-5 \\
%         Epochs & 1 & 1 & 1 \\
%         Optimizer & \multicolumn{3}{c}{AdamW} \\
%         \bottomrule
%     \end{tabular}
%     \caption{Hyperparameters used for BabyLLaVA training across different stages.}
% \end{table}

\paragraph{CVCL Training and Evaluation Details.} We train and evaluate two variants of the CVCL model from \cite{vong2024cvcl} (CVCL-filtered-aug \& CVCL-filtered-random). CVCL-filtered-aug is trained on our filtered SAYCam dataset and our transferred dataset (section \ref{sec:transferred dataset}), both of which contain approximately 67k image-caption pairs. Similarly, CVCL-filtered-random is trained on our filtered SAYCam dataset plus a randomly sampled unprocessed subset of approximately 67k image-caption pairs from LLaVA's pretraining dataset \cite{llava}. We train both variants on a single A100 GPU for 12 hours each using the default hyperparameters specified in the supplemental info of \cite{vong2024cvcl}. For evaluation, we use the model checkpoint from the last training epoch for each variant. 

\clearpage

\paragraph{Out-of-Domain Generalization.} A primary aim of our approach is to ensure baby models align with the cognitive and linguistic limitations of early-stage learners. To empirically validate this property, we explicitly assess baby models on tasks that exceed typical infant-level developmental capacities, such as advanced visual reasoning (Winoground) and general-purpose tasks (VQA and BLiMP). As shown in Table~\ref{tab:main_result_part2}, baby models (e.g., BabyLLaVA, CVCL) perform significantly below upper-bound models, affirming their constrained generalization capabilities. This limitation ensures developmental authenticity, preventing baby models from inadvertently solving complex tasks beyond their intended cognitive stage.

Interestingly, we find that the performance gap between BabyLLaVA and the larger LLaVA-v1.5-7B model is significantly greater on these complex, out-of-domain tasks compared to simpler, in-domain tasks such as VTWT (Table~\ref{tab:main_result_part1}). This indicates that observed differences in performance cannot be attributed solely to differences in model capacity (i.e., parameter count), but also arise from the complexity and alignment of tasks and datasets with the developmental stage being modeled. Thus, baby models' constraints are multidimensional, encompassing not only architectural limitations but also deliberate choices in task design and dataset construction.

\begin{table*}[htbp]
    \centering
    \resizebox{\textwidth}{!}{%
        \begin{tabular}{llccccc}
        \toprule
        \textbf{Category} & \textbf{Model} & \textbf{BLiMP\textsubscript{filtered}} & \textbf{BLiMP\textsubscript{supplement}} & \textbf{Winoground} & \textbf{VQA} & \textbf{DevBench} \\
        \midrule
        \multirow{3}{*}{\textbf{Upper Bound Models}} 
        & LLaVA-v1.5-7B & 0.7299 & 0.8300 & 0.6327 & 0.6273 & 0.8570 \\
        & LLaVA-v1.5-7B-ft & 0.7205 & 0.8032 & 0.5992 & 0.4941 & 0.6300 \\
        & CLIP-large & N/A & N/A & 0.5638 & 0.2397 & 0.7172 \\
        \midrule
        \multirow{2}{*}{\textbf{Baby Models}} 
        & BabyLLaVA-GPT2 & 0.6772 & 0.5903 & 0.5214 & 0.2312 & 0.3907 \\
        & BabyLLaVA-Llama & 0.5095 & 0.5218 & 0.5094 & 0.1477 & 0.3082 \\
        & CVCL & N/A & N/A & 0.5221 & 0.1600 & 0.3993 \\
        \midrule
        \textbf{Random Guess} & - & 0.5000 & 0.5000 & 0.5000 & 0.1250 & 0.3750 \\
        \bottomrule
        \end{tabular}%
    }
    \caption{Evaluation results on out-of-domain benchmarks. For BLiMP, Winoground and VQA, please refer to \cite{choshen2024babylmchallenge} for implementation details. For DevBench, we report the average score of TROG, WG, LWL and VV.}
    \label{tab:main_result_part2}
\end{table*}

\paragraph{Out-of-Domain Tasks Ablation Study.} We also evaluate both our CVCL and BabyLLaVA model variants on several out-of-domain benchmarks, including general purpose benchmarks like VQA, and developmental benchmarks such as DevBench. We make several observations. First, we see that all model variants perform around random chance on Winoground, indicating that none of the models achieve robust compositional reasoning ability. For VQA and DevBench, however, both CVCL and BabyLLaVA variants trained on our transferred dataset (CVCL-filtered-aug \& BabyLLaVA-filtered-aug) achieve superior performance, reinforcing the value of our developmentally adapted general-domain data. 
% However, these gains do not translate to BLiMP\textsubscript{filtered} or BLiMP\textsubscript{supplement} for BabyLLaVA-aug. 
In addition, even the weakest BabyLLaVA model outperform all the CVCL variants on VQA, indicating the advanced reasoning ability of generative VLMs over discriminative VLMs.
Finally, the modest performance across all model variants and tasks reinforces one of our main results that appropriate developmental modeling naturally constrains generalization.

\begin{table}[htbp]
    \centering
    \resizebox{1.0\textwidth}{!}{%
        \begin{tabular}{lccccc} % Adjusted column count
        \toprule
        Model & BLiMP\textsubscript{filtered} & BLiMP\textsubscript{supplement} & Winoground & VQA & DevBench \\
        \midrule
        CVCL-filtered & N/A & N/A & 0.5221 & 0.1600 & 0.3993 \\
        CVCL-filtered-aug & N/A & N/A & 0.4714 & 0.1641 & 0.6086 \\
        CVCL-filtered-random & N/A & N/A & 0.4935 & 0.1173 & 0.5198 \\
        \midrule
        BabyLLaVA-filtered & 0.6772 & 0.5903 & 0.5214 & 0.2312 & 0.3907 \\
        BabyLLaVA-filtered-aug & 0.6646 & 0.5061 & 0.5455 & 0.4064 & 0.5303 \\
        BabyLLaVA-filtered-random & 0.6746 & 0.4778 & 0.5335 & 0.3659 & 0.4722 \\
        \bottomrule
        \end{tabular}%
    } % Closing brace correctly placed
    \caption{Ablation study results on out-of-domain benchmarks. For BLiMP, Winoground and VQA, please refer to \cite{choshen2024babylmchallenge} for implementation details and metrics. For DevBench, we report the average score of TROG, WG, LWL and VV.}
    \label{tab:dataset_ablation_part2}
\end{table}

\begin{figure}[htbp]
  \centering
  % \fbox{\rule{0pt}{0.5in} \rule{0.9\linewidth}{0pt}}
  \includegraphics[width=1.0\linewidth]{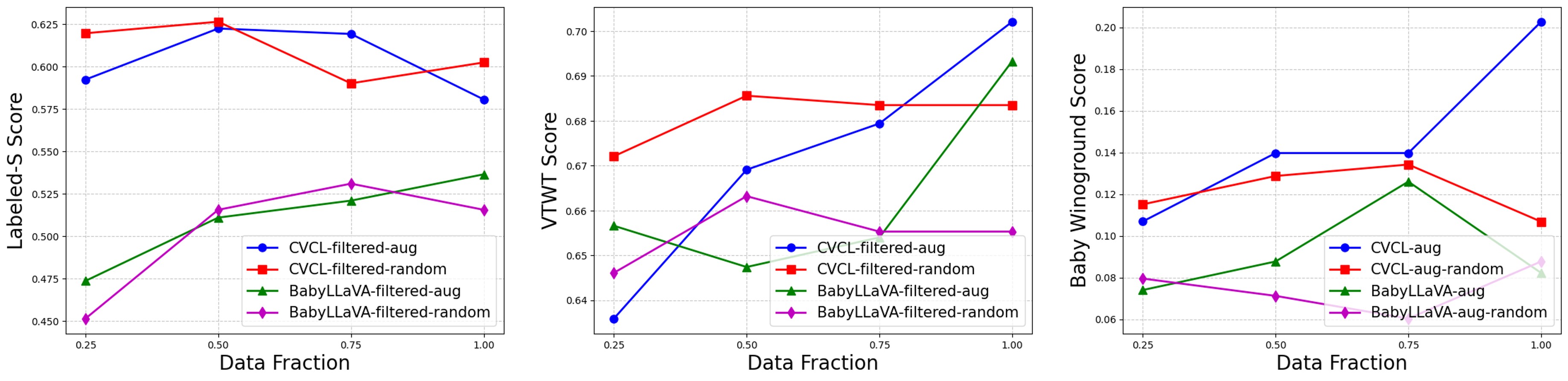}
   \caption{Performance on different fractions of datasets.}
   \label{fig:ratio}
\end{figure}

\paragraph{Transferred Data Efficiency Ablation Study.} To further investigate the extra data efficiency brought by our introduced transferred dataset, we subsample \textit{filtered-aug} and \textit{filtered-random} datasets by different fractions, then perform model training with identical training steps, and test on our in-domain benchmarks. Figure \ref{fig:ratio} shows the result. The \textit{filtered-aug} curve rises more steeply, and using only 25–50\% of \textit{filtered-aug} already equals the full \textit{filtered-random} on in-domain tasks, confirming substantial sample-efficiency gains.

\end{document}